\def\year{2018}\relax
\documentclass[letterpaper]{article} 
\usepackage{aaai18}  
\usepackage{times}  
\usepackage{helvet}  
\usepackage{courier}  
\usepackage{url}  
\usepackage{graphicx}  
\usepackage{CJKutf8}
\usepackage{latexsym}
\usepackage{multirow}
\usepackage{amsmath}
\usepackage{booktabs}
\usepackage{subfigure}

\begin{document}
%
\title{Towards Automatic Generation of Entertaining Dialogues in Chinese Crosstalks}
\author{Shikang Du, Xiaojun Wan, Yajie Ye\\
Institute of Computer Science and Technology, The MOE Key Laboratory of Computational Linguistics\\
Peking University, Beijing 100871, China\\
\{dusk, wanxiaojun, yeyajie\}@pku.edu.cn\\
}

\begin{CJK*}{UTF8}{gbsn}
	
\maketitle
\begin{abstract}
	Crosstalk, also known by its Chinese name {\it xiangsheng}, is a traditional Chinese comedic performing art featuring jokes and funny dialogues, and one of China's most popular cultural elements. It is typically in the form of a dialogue between two performers for the purpose of bringing laughter to the audience, with one person acting as the leading comedian and the other as the supporting role. Though general dialogue generation has been widely explored in previous studies, it is unknown whether such entertaining dialogues can be automatically generated or not. In this paper, we for the first time investigate the possibility of automatic generation of entertaining dialogues in Chinese crosstalks. Given the utterance of the leading comedian in each dialogue, our task aims to generate the replying utterance of the supporting role. We propose a humor-enhanced translation model to address this task and human evaluation results demonstrate the efficacy of our proposed model. The feasibility of automatic entertaining dialogue generation is also verified.
	
\end{abstract}

\section{Introduction}

Crosstalk, also known by its Chinese name \textit{相声/xiangsheng}, is a traditional Chinese comedic performing art, and one of China's most popular cultural elements. It is typically in the form of a dialogue between two performers, but much less often can also be a monologue by a solo performer, or even less frequently, a group act by multiple performers. The crosstalk language, rich in puns and allusions, is delivered in a rapid, bantering style. The purpose of Xiangsheng is to bring laughter to the audience, and the crosstalk language features humorous dialogues \cite{link1979genie,moser1990reflexivity,terence2013china,mackerras2013performing}.  

The language style of crosstalk is just like chatting or gossip, but is more funny and humorous, especially in crosstalks given by two performers. It would be an ideal resource for studying humor in dialogue system.

However, there are some special rules in crosstalks. 
For the crosstalk between two performers, one person acts as the leading comedian (or {\it 逗哏/dougen} in Chinese) and the other as the supporting role (or {\it 捧哏/penggen}). The two performers usually stand before an audience and  deliver their lines in rapid fire by turn. They echo each other in the crosstalk performance. In each turn, the leading role usually tells stories and jokes, or does some sound imitation in his utterance, and the supporting role points out the humorous point in the leading role's performance, or even adds fuel to the leading role's performance, making it funnier. For example, 

\begin{tabular}{ll}
	\textbf{A}: & 楚国大夫屈原，五月初五死的，我们\\
	&	应该永远怀念屈原。要是没有屈原，\\
	&	我们怎么能有这三天假期呢？ \\
	&	The mid-autumn festival is in memory of Qu \\
	&	Yuan. We should keep him in mind forever, \\
	&	because his death brings us this 3-day holiday.\\
	\textbf{B}: & 这个，代价大点儿。\\
	&	It costs him a lot (to have a holiday).\\
	\textbf{A}: & 我觉得应该再多放几天假。\\
	&	I think it would be better with more holidays.\\
	\textbf{B}: & 那得死多少人啊。\\
	&	How many people would die then!\\
\end{tabular}
In this example, B acts as the supporting role. His last response unexpectedly links the number of holiday with the number of people died, which makes the whole dialogue more funny.
But in many cases, the supporting one acts as a go-between, gives positive response (such as ``当然/Of course'' or ``这样/That's why'') or negative response (such as ``啊？/Ah?''), and sometimes repeats key points in the leading role's utterance, making the narration given by the leading role go smoothly (e.g. A: 虽然道路崎岖，所幸还有蒙蒙月色/ Although the road is rough, the moonlight is bright. B:还能看见点/ We can still see things on the road.) In brief, the crosstalk between two performers can be considered a special and challenging dialogue form - the entertaining dialogue.      

Though general dialogue generation has been widely explored and achieved great success in previous studies \cite{li-EtAl:2016:EMNLP20162,sordoni-EtAl:2015:NAACL-HLT,ritter2011data}, 
it is unknown whether such entertaining dialogues can be automatically generated or not. If computers can generate entertaining dialogues well, the AI ability of computer will be further validated. The function of generating entertaining dialogues is also very useful in many interactive products, making them more appealing. In this study, we for the first time investigate the possibility of automatic generation of entertaining dialogues in Chinese crosstalks. Given the utterance of the leading comedian in each dialogue, our task aims to generate the replying words of the supporting role. 

We propose a humor-enhanced translation model to address this special and challenging task, and the model explicitly leverages a sub-model to measure the humorous characteristic of a dialogue. Human evaluation results on a real Chinese crosstalk dataset demonstrate the efficacy of our proposed model, which can outperform several retrieval based and generation based baselines. The feasibility of automatic entertaining dialogue generation is also verified. 

The contributions of this paper are summarized as follows:

1) We are the first to investigate the new task of entertaining dialogue generation in Chinese crosstalks. 

2) We propose a humor-enhanced translation model to address this challenging task by making use of a sub-model to measure the humorous characteristic of a dialogue.

3) Manual evaluation is performed to verify the efficacy of our proposed model and the feasibility of automatic entertaining dialogue generation.

In the rest of this paper, we will first describe the details of our proposed model and then present and discuss the evaluation results. After that, we introduce the related work. Lastly, we conclude this paper. 

\section{Our Generation Method}

Given an utterance  $\mathbf{s}$ of the leading role (i.e. \textit{dougen}) in Chinese crosstalks, our task aims to generate the replying utterance $\mathbf{r}$ of the supporting role (i.e. {\it penggen}), which is called crosstalk response generation (CRG). The generated utterance needs to be fluent and related to the leading role's utterance. Moreover, it is also expected that the generated utterance can make the dialogue more funny and entertaining.  

As mentioned earlier, our task is a special form of dialogue generation. In recent years, there are many methods proposed for dialogue generation based on a large set of training data, including the deep learning methods (especially sequence-to-sequence models) \cite{li-EtAl:2016:EMNLP20162}. However, deep learning methods usually require a large training set to achieve good performance in dialogue generation tasks, which is hard to obtain for our task. So, we choose a more traditional but effective way based on machine translation to address the new task of crosstalk response generation.

\begin{figure}[hbt]
	\includegraphics[width=80mm]{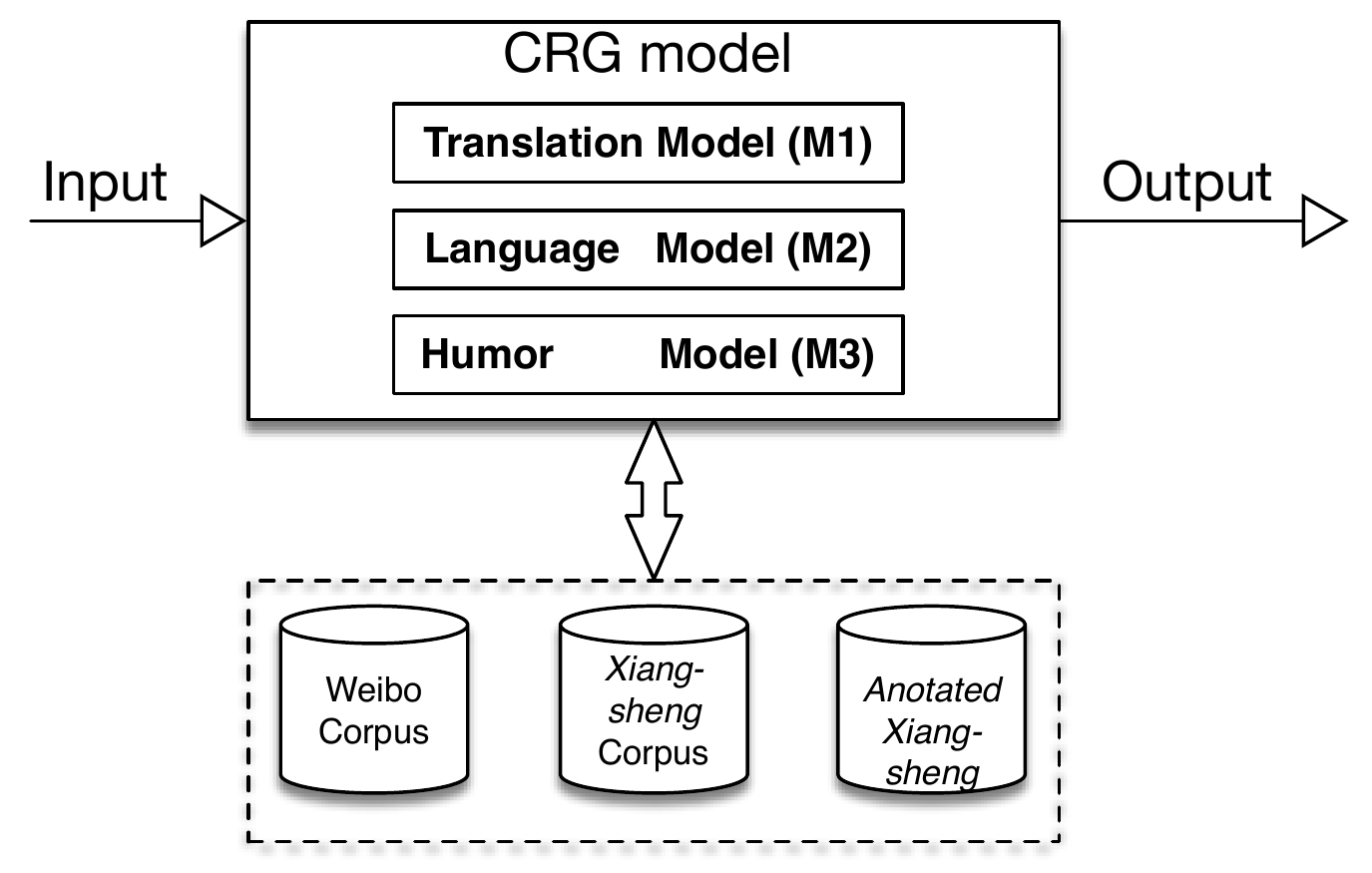}
	\caption{\label{fig:sys} General architecture of our system}
\end{figure}

\textit{penggen} often gives comments on \textit{dougen}'s utterance, sometimes \textit{penggen} even retells the \textit{dougen}'s words but in a more humorous way. We believe that the \textit{dougen}'s response has some potential patterns according to
the utterance given by \textit{penggen}, and treat response generation as a monolingual translation problem, in which the given input (utterance given by \textit{dougen}) is treated as the foreign language and the humorous response as the source language. 
Machine translation (MT) has already been successfully used in response generation \cite{ritter2011data}, in which input post was seen as a sequence of words, and word or phrase based translation was made to generate another sequence of words as response. If we simply treat crosstalk response generation as a general dialogue generation problem, we can apply statistical machine translation (SMT) model \cite{koehn2003statistical} to generate responses accordingly, ignoring the entertaining characteristic of crosstalk. 
In machine translation, beam search is used in decoding process, which could generate multiple candidates with scores. Usually only the candidate with the highest score could be accepted. These scores reflects the similarity of the candidate and reference.
However, just like that some question may have many different answers, there might still be acceptable, or even unexpected but wonderful candidates with lower scores. It's a pity to get these good response ignored just because they shares little similarity with the references in a limited training dataset. 
To exploit them, and also to address the crosstalk generation problem, we propose a humor-enhanced machine translation model to generate response utterance in crosstalk. Our proposed model leverages a sub-model to explicitly model the degree of humor of a dialogue, and integrate it with other sub-models, as illustrated in \textbf{Figure \ref{fig:sys}}. 

\subsection{Response Generation Model}

We get pairs of aligned utterance and response from the dialogue fragments in Chinese crosstalks, which are considered monolingual parallel data. The two performers echo each other in a crosstalk, and their roles keep consistent in the whole crosstalk,  and the leading role and the supporting role of each utterance can be easily identified. Then we segment the utterances into words. Each pair consists of a sequence of words $\mathbf{s} (\{s_1, s_2, ..., s_l\})$ spoken by the leading role, and a sequence of words $\mathbf{ref}$ replied by the supporting role, while the response we generated is denoted as $\mathbf{r} (\{r_1, r_2, ..., r_l\})$.

Given the leading role's utterance $\mathbf{s}$, we aim to generate the best response utterance $\mathbf{r}$ by using our proposed generation model. The proposed generation model has three sub-models(M1, M2, M3): translation model, language model and humor model. We will introduce each sub-model and then introduce the framework of model combination. 

\subsubsection{Translation Model (M1)}

The translation model translates the given leading role's utterance $\mathbf{s}$ into a sequence of words $\mathbf{r}$, which is treated as the response. Let $({s}_i, {r}_i)$ be a pair of translation units, we can compute the word translation probability distribution $\phi_{tm}({s}_i, {r}_i)$ , which is defined in \cite{koehn2003statistical}. 
Each word $s_i$ in input utterance is translated to a word in response $r_i$, and the word in response would be reordered. 

Reordering of generated response is modeled by 
a relative distortion probability distribution $d(a_i - b_{i-1})$, where $a_i$ is the starting position of the word in the input utterance $\mathbf{s}$ translated to the $i$-th word in generated response $\mathbf{r}$, and $b_{i-1}$ denotes the end position of the word in the input utterance translated into the $(i-1)$-th word in the response. We use $d=\alpha^{|x-1|}$ as implementation.


Thus, the translation score between the leading role's utterance $\mathbf{s}$ and generated response $\mathbf{r}$ is:
\begin{equation}
	p_{tm}\left( \mathbf{r}, \mathbf{s} \right) = \prod_{i=1}^{l}\phi_{tm}({s}_i, {r}_i) d(a_i - b_{i-1})
\end{equation}

\subsubsection{Language Model (M2)}

We use a 4-gram language model in this work. The language model based score is computed as:

\begin{equation}
	p_{lm}({\mathbf r}) = \prod_j p(r_j|r_{j-3}r_{j-2}r_{j-1})
\end{equation}
where $r_j$ is the $j$-th element of $\mathbf r$.

%
%
%

\subsubsection{Humor Model (M3)}

We want to build a model to measure the degree of humor of a dialogue. However, humor is very complex. In Chinese crosstalks, humor can be expressed by the actors' tone, body language and verbal language. In this study, we mainly focus on modeling the verbally expressed humor in crosstalks.

We build a classifier to determine the probability of being humorous for each response candidate in the context of the input utterance. In this model, we evaluate humor in 4 dimensions, just as the same as \cite{yang2015humor} : (a) Incongruity, (b) Ambiguity, (c) Interpersonal Effect, and (d) Phonetic Style.

Incongruity structure plays an important role in verbal humor, as stated in \cite{lefcourt2001humor} and \cite{paulos2008mathematics}. Although it is hard to determine incongruity, it is relatively easier to calculate the semantic disconnection in a sentence. We use Word2vec to derive the word embeddings and then compute the distances between word vectors.

When a listener expects one meaning, but is forced to use another meaning \cite{yang2015humor}, there is ambiguity. This distraction often makes people laugh. To measure the ambiguity in the sentence, we collect a number of antonyms and synonyms for feature extraction. Note that antonyms are used as as an important feature in humor detection in \cite{mihalcea2005making}. Using Chinese WordNet \cite{huang2010infrastructure}, we get the pairs of antonyms and synonyms.

Interpersonal effect is associated with sentimental effect \cite{zhang2014recognizing}. A word with sentimental polarity reflects the emotion expressed by the writer. We use a dictionary in \cite{Xu2008} to compute the sentimental polarity of each word, and add them up as the overall sentimental polarity of a sentence.

Many humorous texts play with sounds, creating incongruous sounds or words. Homophonic words have more potential to be phonetically funny. We count the number of homophonic words and words with the same rhyme, with the help of pypinyin\footnote{http://pypi.python.org/pypi/pypinyin} .

Furthermore, adult slang is described in \cite{mihalcea2005making} as a key feature to recognize jokes, so we count the number of slangs.

Note that we extract features from the response alone and also extract features from the whole turn of dialogue consisting of both the given input utterance and the response. To summarize, the features we use are listed below:

\begin{itemize}
	\item minimum and maximum distances of each pair of word vectors in the response;
	\item minimum and maximum distances of each pair of word vectors in the whole turn of dialogue (including the given input utterance and the response);
	\item number of pairs of antonyms in the response;
	\item number of pairs of antonyms in the whole turn of dialogue;
	\item number of pairs of synonyms in the response;
	\item number of pairs of synonyms in the whole turn of dialogue;
	\item sentimental polarity in the response;
	\item sentimental polarity in the whole turn of dialogue;
	\item number of homophonic words in the response;
	\item number of homophonic words in the whole turn of dialogue;
	\item number of the words with same rhyme in the response;
	\item number of the words with same rhyme in the whole turn of dialogue;
	\item number of slangs in the response.
\end{itemize}

We choose the random forest classifier \cite{Liaw2002} because it generally outperforms other classifiers based on our empirical analysis. 

The output probability for $\mathbf{r}$ is used as the humor model score $p_{hm}({\mathbf r})$.

\subsubsection{Model Combination}
We use a log-linear framework to combine the above three sub-models and get our response generation model. Note that the translation model corresponds to two parts. 

\begin{equation}
	\begin{split}
		p({\mathbf r}|{\mathbf s})
		&= \lambda_{tm} \sum_i \log \phi_{tm}(s_i, r_i)\\
		&+ \lambda_{ds} \sum_i \log d(a_i - b_{i-1})\\
		&+ \lambda_{lm} \sum_j \log p(r_j|r_{j-3}r_{j-2}r_{j-1})\\
		&
		+ \lambda_{hm} \log p_{hm}({\mathbf r})
	\end{split}
\end{equation}
where $\lambda_{tm}$, $\lambda_{ds}$, $\lambda_{lm}$
and $\lambda_{hm}$ are weight parameters of the sub-models and can be learned automatically.  

\subsection{Learning and Decoding}
In the model M1, we use relative frequency to estimate the word translation probability distribution $\phi_{tm}({s}_i, {r}_i)$, and no smoothing is performed. 

\begin{equation}
	\phi_{tm}(s, r) = \frac{\mathrm{count}(s,r)}{\sum_s (s,r)}
\end{equation}
A special token NULL is added to each utterance and aligned to each unaligned foreign word.

The training process is similar to that in \cite{ritter2011data}.
We use the widely used toolkit Moses \cite{koehn2007moses} to train the translation model.

The scikit-learn toolkit\footnote{http://scikit-learn.org/stable/index.html} is used and the probability of prediction is acquired through the API function of {\tt predict\_proba}.

In order to estimate weight parameters in the combined model, we apply the minimum error rate training (MERT) algorithm \cite{Och:2003:MER:1075096.1075117}, which has been broadly used in SMT.  The most common optimization objective function is BLEU-4
\cite{papineni2002bleu}, which requires human references. We take the original human response derived from our parallel corpus as the single reference. We use the tool Z-MERT \cite{zaidan2009z} for estimation. The weight parameter values that lead to the highest BLEU-4 scores on the development set are finally selected.

In the decoding process, we use the beam search algorithm to generate the top-100 best 
response candidates for each input utterance based on M1 and M2. Then we obtain the score of M3 of each candidate, rank the candidates according to the combined model and select the best candidate as output.

\subsection{Final Reranking with the Humor Model}

Note that the above combined model is optimized for the BLEU-4 score, but the BLEU-4 score cannot well reflect the humorous aspect of generated responses, so in order to improve the humor level of a dialogue, we further select the top-five best response candidates generated by the above combined model and rerank them according to the score of the humor model (M3), and finally use the top-ranked one as the output. Note that We use only top five candidates in this step because it is more efficient and effective to rerank a small number of high-quality candidates, while the readability and relevance of other candidates with low ranks cannot be guaranteed. The number of five is determined based on the development set.    

\section{Experiment Setup}

\subsection{Experiment Data}

We collect the crosstalk data from multiple sources: (a) published books\footnote{(1)Liu Yingnan. {\it A Complete Collection of China Traditional Cross Talks 5 Vols.}, Culture and Art Publishing House, 2010.~~ (2)Wang Wenzhang. {\it Famous Crosstalk Actor's Masterpiece Series},  Culture and Art Publishing House, 2004. et,al.}; (b) websites\footnote{(1) http://www.xiangsheng.org; (2) http://www.tquyi. com; et, al. 
} , where Chinese crosstalk fans collect and collate existing famous crosstalk masterpieces.  (c) records of crosstalk play.
The dataset we collect consists of over $173,000$ pairs of utterances, from $1,551$ famous excerpts of crosstalks. Since long sentences would slow down our training process, we filtered out responses longer than 60 words. In order to improve qualty, we also filtered  out very short responses that are usually 1 modal particles. Over $150,000$ utterances was used in our dataset after this process.

We divide the pairs of utterances and responses in the dataset into three parts, and we randomly select $2000$ pairs as the test set, $4000$ pairs as the development set for weight parameter estimation, and the rest as the training set for translation model.

Since training language model requires a large-scale dataset, which could hardly be offered in the domain of crosstalks, we add Chinese microblog messages from Sina Weibo\footnote{http://weibo.com} to enlarge the corpus for language model training. The language styles in Weibo and Chinese crosstalks are quite similar in that the sentences in Weibo messages and crosstalks are usually short and informal. We collect 6 million pieces of Weibo messages and comments from Sina Weibo.

Not all utterances in Chinese crosstalks are humorous, because there are many utterances serving as go-betweens, so we have to manually build the training data for humor model learning. Because of the lack of Chinese humorousness dataset, we randomly collect $6000$ pairs of utterances in Chinese crosstalks, and manually label them into two classes: {\it humorous} or {\it not humorous}. $348$ pairs are marked as  {\it humorous}, and we replicate the minor class instances and remove some major class instances to make the class distribution more balanced. 

Then we use the labeled data for training the random forest classifier in the humor model.

\subsection{Comparison Methods}

We implement retrieval based methods for comparison:
\begin{itemize}
	\item {\sc \textbf{Seq2Seq}}: Treat this problem as translation problem with \textsc{Seq2Seq} model with attention. GRU cells  are used in RNN, and number of cells are 256.
	\item {\sc \textbf{Ir-Ur}}: Retrieve the response which is most similar to the input utterance from both the development set and the training set.
	\item {\sc \textbf{Ir-Uu}}: Retrieve the most similar utterance to the input utterance, and then return the response associated with the retrieved utterance;
	\item {\sc \textbf{Ir-Cxt}}: Retrieve the response which is most similar to the input utterance and three previous utterances of the input utterance;
\end{itemize}

Similarity was calculated by comparing word-level cosine similarity.

Our proposed method consists of all the three sub-models (including the final selection step), named as {\sc \textbf{Smt-H}}. We further compare our method with the basic machine translation method considering two sub-models M1, M2, named as {\sc \textbf{Smt}}. 

Note that in our method, the humor model is used in both the combined model and the final reranking process. 

\subsection{Evaluation Metrics}

We adopt human evaluation to verify the effectiveness of our system. We also report automatic evaluation results with BLEU \cite{papineni2002bleu}. But in the dataset only one reference response is provided for each given utterance, and the humor aspect cannot be well captured by the BLEU metrics. 
So we rely both on automatic and human evaluation results for this special task. 

We employ two human judges to rate each generated response in three aspects:

\textbf{Readability}: It reflects the grammar correctness and fluency;

\textbf{Entertainment}: It reflects the level of humor of the response;

\textbf{Relevance}: It reflects the semantic relevance between the input utterance and the response generated. It also reflects logic and sentimental consistency.

Each judge is asked to assign an integer score in the range of 0 $\sim$ 2 to each generated response with respect to each aspect. The score $0$ means ``poor'' or ``not at all'', $2$ means ``good'' or ``very well'', and $1$ means ``partially good'' or ``acceptable''. For example, In readability, $1$ means that there are some grammar mistakes but human evaluator can still understand the meaning of the response. 




To help human raters to determine whether the generated response is relevant to the input utterance, we also provide previous two rounds of dialogues of the input utterance to the raters. 

\section{Result and Analysis}
\subsection{Automatic Evaluation Results}

As shown in \ref{res-auto}, we found the BLEU score of our \textsc{Smt-H} is higher than baselines in the $2000$ utterances test set, which can be simply explained since more global features could be accessed in our \textsc{Smt-H} model. 
	\textit{t}-test results on BLEU scores of the two model show that their difference is significant ($p< 1\times 10{-5}$).

\begin{table}[htb]
\begin{tabular}{lllll}
\toprule
 & BLEU-4 & BLEU-3 & BLEU-2 & BLEU-1 \\
 \midrule
\textsc{Smt-H} & \textbf{16.62} & \textbf{18.99} & \textbf{22.41} & \textbf{29.57} \\
\textsc{Smt} & 15.13 & 17.39 & 20.62 & 27.39 \\
\textsc{Seq2Seq} & 16.03 & 17.76 & 20.63 & 27.66 \\
\textsc{Ir-Uu} & 2.6 & 3.51 & 5.52 & 12.53 \\
\textsc{Ir-Ur} & 4.4 & 5.15 & 6.83 & 13.13 \\
\textsc{Ir-Cxt} & 3.14 & 4.17 & 6.34 & 13.76 \\
\textsc{Rnd} & 0.00 & 0.00 & 1.65 & 9.73 \\
\bottomrule
\end{tabular}
	\caption{Automatic Evaluation Result of \textsc{Smt}, \textsc{Smt-H} and \textsc{Seq2Seq}}
	\label{res-auto}
\end{table} 
As expected, the \textsc{Seq2Seq} model could get better scores in our automatic test than ordinary \textsc{Smt} model, but not better than our \textsc{Smt-H} model. A larger training set might help improve performance of the deep learning based model.
	
BLEU scores of all \textsc{Ir} based models are lower than $5\%$ 
 One possible reason is that the crosstalk dataset is not very large, and the utterances in the dataset are very diversified, so it is hard for retrieval based methods to find proper responses from the dataset directly. While generation based methods are more flexible and they can generate new responses for input utterances. For the retrieval based methods, IR-UR performs better than IR-UU, which is contrary to our intuition. This phenomenon has been discussed in \cite{ritter2011data}.


\subsection{Human Evaluation Results}

\begin{figure*}
	\centering
	\subfigure[Readability scores]{
		\label{fig:score:read}
		\includegraphics[height=1.6in]{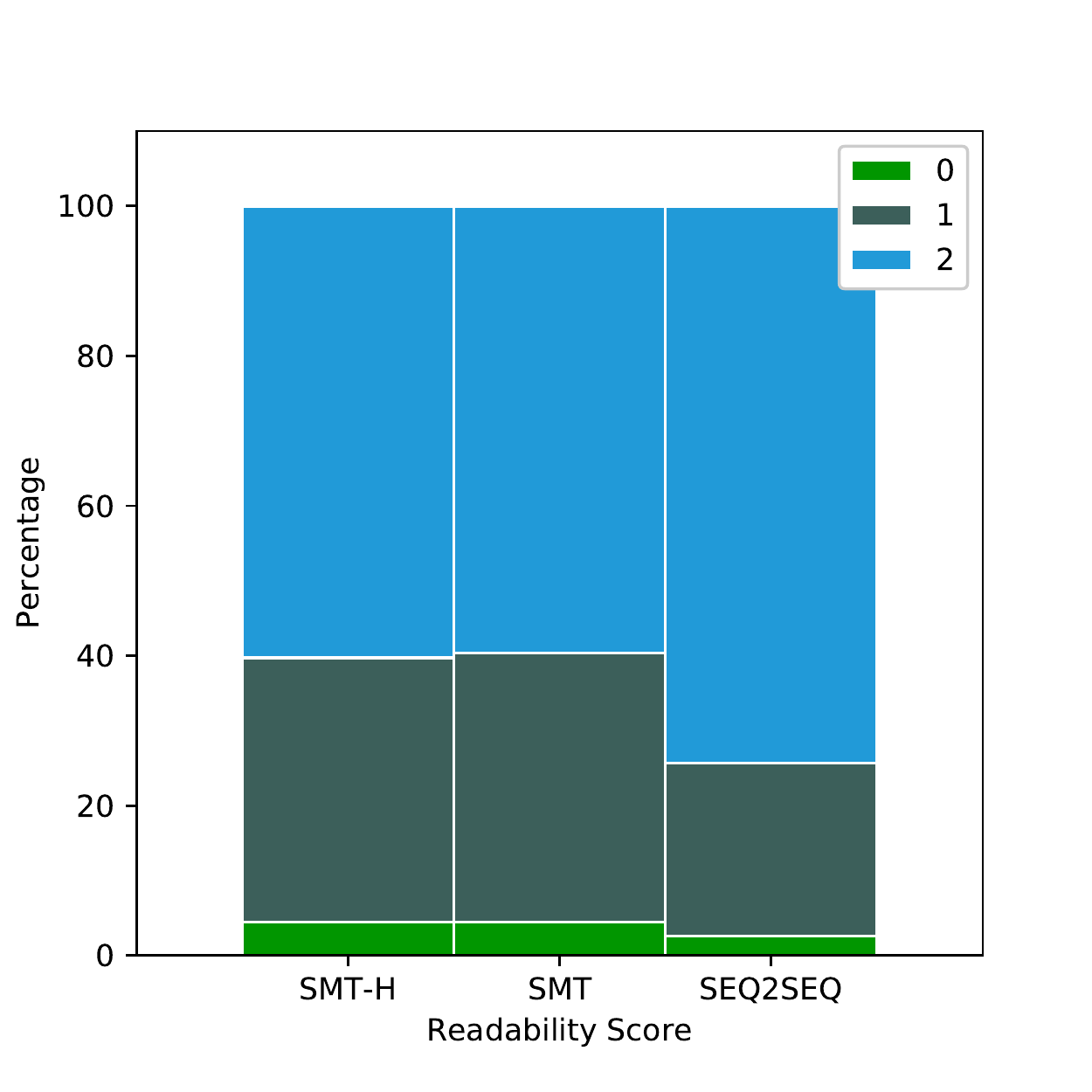}
	}
	\subfigure[Entertainment scores]{
		\label{fig:score:ente}
		\includegraphics[height=1.6in]{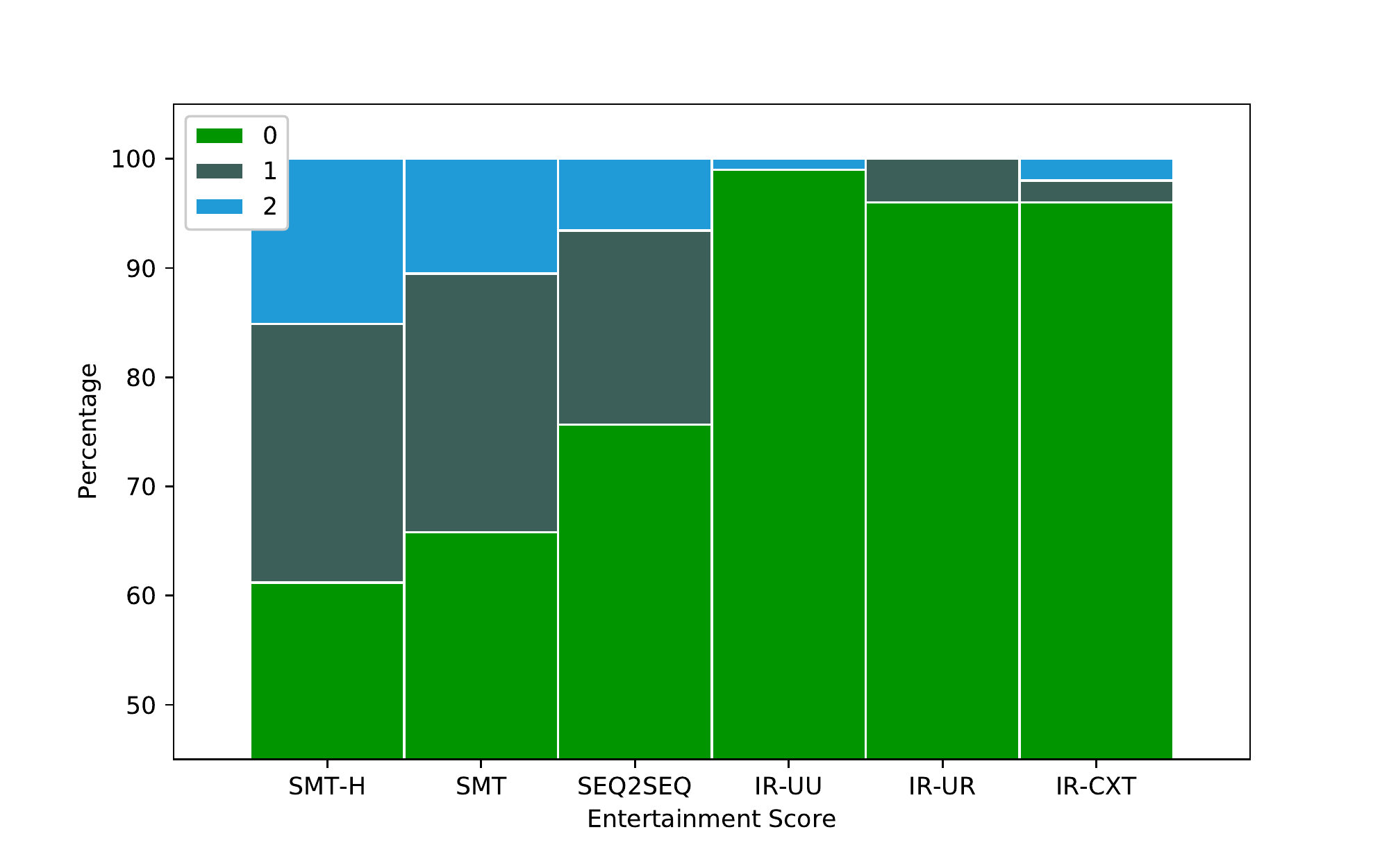}
	}
	\subfigure[Relevance scores]{
		\label{fig:score:rele}
		\includegraphics[height=1.6in]{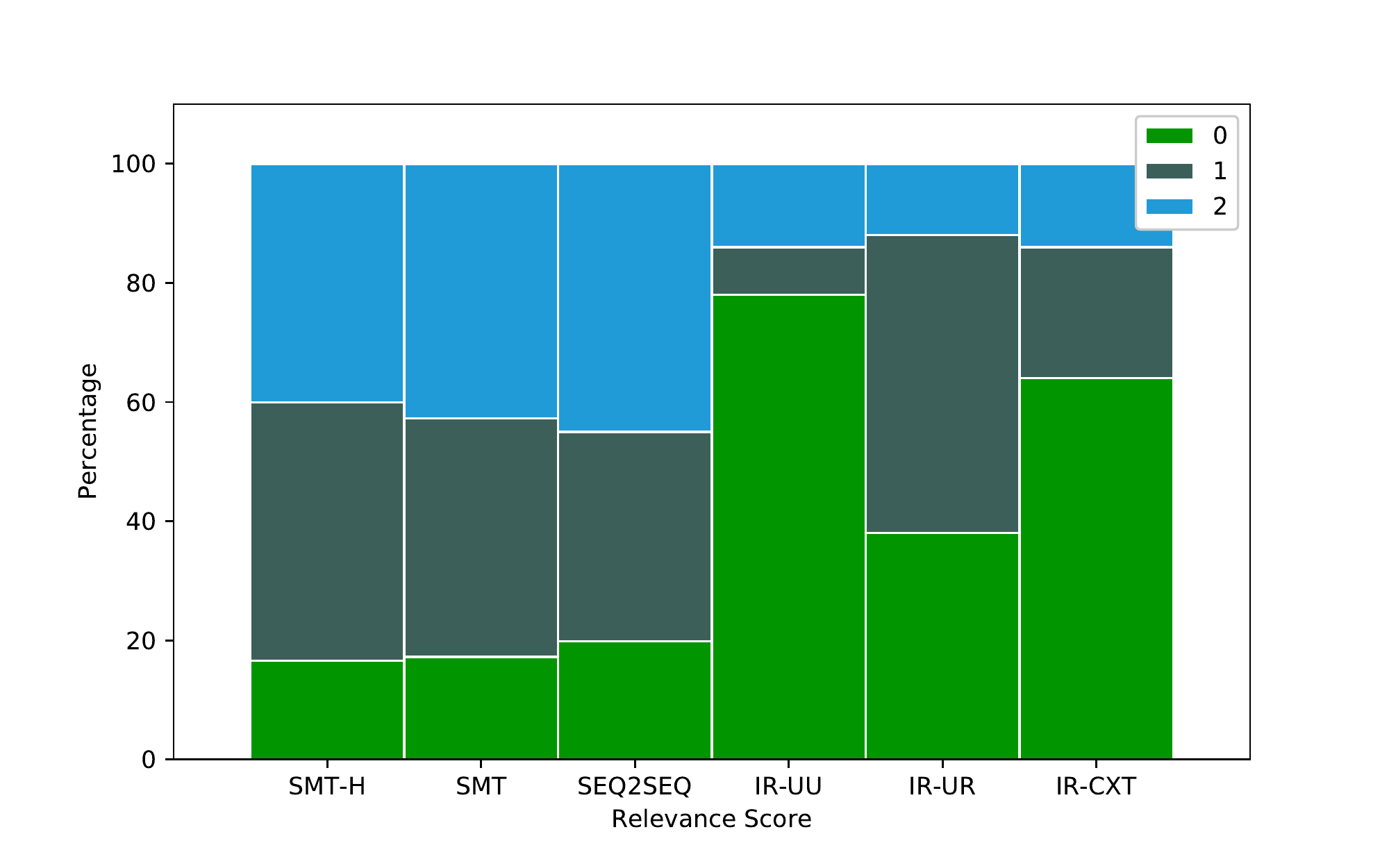}
	}
	\caption{Percentages of human evaluation scores}
	\label{fig:score}
\end{figure*}

We randomly selected $150$ input utterances in the test set and asked two raters to label the responses generated by retrieval based methods and machine translation based methods. 
The percentage of each rating level is calculated for each method with respect to each aspect, as shown in \textbf{Figure \ref{fig:score}}. 
Since retrieval based methods extract existing utterances directly from the dataset, the readability of the retrieved responses is usually very good and thus we do not need to label the readability of these responses. We further compute the relative ratio of the average rating score of each method to the average rating score of the basic {\sc Smt} model with respect to each aspect, as shown in \textbf{Table \ref{res-relative}}, and a ratio score larger than 100\% means the corresponding method performs better than the basic {\sc Smt} model, while a ratio score lower than 100\% means the corresponding method performs worse than the basic {\sc Smt} model.  

\begin{table*}[tb]
	\centering
	\begin{tabular}{lllllll}
		\toprule
		& \textsc{Smt} & \textsc{Smt-H} & \textsc{Seq2Seq} & \textsc{Ir-Uu} & \textsc{Ir-Ur} & \textsc{Ir-Cxt} \\
		\midrule
		Readability & 100.00\% & 100.41\% & 110.74\% & - & - & - \\
		Entertainment & 100.00\% & 120.59\% & 69.12\% & 4.47\% & 8.94\% & 13.41\% \\
		Relevance & 100.00\% & 98.42\% & 99.74\% & 28.69\% & 58.97\% & 39.84\% \\
		\bottomrule
	\end{tabular}
	\caption{Relative ratios of average rating scores of each method to that of \textsc{Smt}}
	\label{res-relative}
\end{table*} 

As can be seen from the human evaluation results, the responses generated by machine translation based methods are much more entertaining and relevant than retrieval based methods. We also find that that {\sc Ir-Ur} returns more relevant responses than {\sc Ir-Uu}. 
It also reveal that the translation model could generate more entertaining but less fluent response than \textsc{Seq2Seq} model. It could be explained that the responses generated by \textsc{Seq2Seq} model are too ordinary to get people feel amused.

Comparing {\sc Smt} with {\sc Smt-H}, we can see that {\sc Smt-H} receives higher rating scores than {\sc Smt} with respect to fluency and entertainment. The comparison results demonstrate that the use of the humor model can indeed make the generated responses more entertaining, which is very important for Chinese crosstalks. An auxiliary effect by using the humor model is to improve the readability of the generated responses. 


Now we show two examples of input utterances and different responses generated by {\sc Ir-Uu}, {\sc Smt} and {\sc Smt-H} as follows:

\noindent
\begin{tabular}{ll}
	\toprule
	\textbf{A}: & 躲？我把扁担一横，立托千斤\\
	\textit{(context)}& 闸，迎着他的铁棍——咔！ /Hide?\\
	& I bravely used my shoulder pole to\\
	& meet against his iron bar... Crack!  \\
	\textbf{B}: & 铁棍飞 。/His iron bar dropped?\\
	\midrule
	\textbf{A(input)}: & 扁担折了。  /No, my shoulder pole \\
	&  was fractured. \\
	\midrule
%
	\textbf{B(\textsc{Smt})}: & 搁在一块儿说个绕嘴的折了拿\\
	& 什么打？ /Put them together  \\
	& and tell me a tongue twister,  \\
	& what else could be used to meet \\
	& against his weapon?\\
	\textbf{B(\textsc{Smt-H})}: & 好 --- 折啦 ？ \\  
	&  /What? It was fractured?\\
	\textbf{B(\textsc{Seq2Seq})}: & 这回该二朝的关系? /\\
	& (\textit{Broken sentences})\\
	\textbf{B(\textsc{Ir-Uu})}: & 都知道。\\
	& /We all know that.\\
	\bottomrule
	\\
\end{tabular}

\noindent
\begin{tabular}{ll}
	\toprule
	\textbf{B}\textit{(context)}: & 老乡们 / My Fellow villagers\\
	\midrule
	\textbf{A(input)}: & 猪尾巴 / The pork tail\\
	\midrule
	\textbf{B(\textsc{Smt})}: & 噢！才生下来的？你？\\
	& /Were just born? You? Are you \\
	& kidding me?\\
	\textbf{B(\textsc{Smt-H})}: & 注意吧 / Pay attention\\  
	\textbf{B(\textsc{Seq2Seq})}: & 注意吧 / Pay attention\\ 
	\textbf{B(\textsc{Ir-Uu})}: & 嗬 / Ho\\
	\bottomrule
\end{tabular}

In the first example, the response generated by \textsc{Smt-H} is more related to the input utterance. In the second example, the Chinese phrase 猪尾巴/``pork tail'' has the same pronunciation with 注意吧/``pay attention'' in the response generated by \textsc{Smt-H}. It is laughable since the supporting role (B) distorts A's utterance's meaning with a skill of homophonic, while the responses generated by \textsc{Smt} and \textsc{Ir-Uu} are totally irrelevant. 

\subsection{Discussion}

Our method works well with short input. It can generate appropriate responses which act as go-between in narration. For example,

\textbf{A}: 你爸爸穿衣裳也讲究。/ Your father is dainty about his dress. 

\textbf{B}(\textsc{Smt-H}): 怎么讲究？/ Dainty? How ?

Some responses generated by our method are entertaining. For example,

\textbf{A}: 龙生龙, 凤生凤, 老鼠的儿子会打洞/Dragon born dragon, chicken born chicken, mouse's son could only make hole. (\textit{You are just like a mouse.}) 

\textbf{B}(\textsc{Smt-H}): 你不是后继无鼠了吗这代, 您这套我都会了。/ You even don't have a mouse-like successor! I've found out your strategy.

%
%
%
%

With the use of the humor model in the combined model and the use of it in the final reranking process, our method can generate better response. For example,

\textbf{A}: 比如说我是天上的一颗星星。/ For example, I am a star in the heaven.

\textbf{B}(\textsc{Smt-H}): 噢这是不是。 / Oh, is it this one?

\textbf{B}(\textsc{Smt}): 是噢是你。 / Well, it's you.

\textbf{B}(\textsc{Seq2Seq}): 可不对。 / That's right.

%
%
%
%

However, there are still several shortcomings for our method:

1) Some generated responses are not fluent and the readability is not good. Some responses are broken sentences. For example,

\textbf{A}: 好，新春进步！/Well, hope you make progress in new spring. 

\textbf{B}(\textsc{Smt-H}): 不春进步了。/Not spring progress. 

The reason may be that the current crosstalk corpus is not adequate for training a high-quality language model, but unfortunately it is hard to obtain a large crosstalk corpus because fewer and fewer people still work on this performing art and create new crosstalks.   

2) In some cases,  our method will only give the input words back without translation and rewriting (e.g. 八匹马呀/Ah, there are eight horses). This may be caused by the data sparsity problem in the dataset. If the words or expressions do not or seldom appear in the training corpus, our method cannot find any "translations" to them and can only return them back directly.  


\section{Related Work}

The most closely related work is dialogue generation Previous work in this field relies on rule-based methods, from learning generation rules from a set of authored labels or rules \cite{Oh:2000:SLG:1117562.1117568,banchs2012iris}
to building statistical models based on templates or heuristic rules \cite{levin2000stochastic,pieraccini2009we}.  
li-EtAl:2017:EMNLP20175
After the explosive growth of social networks, the large amount of conversation data enables the data-driven approach to generate dialogue. Research on statistical dialogue systems fall into two categories: 1) information retrieval (IR) based methods \cite{ji2014information}, 2) the statistical machine translation (SMT) based methods \cite{ritter2011data}. IR based methods aim to pick up suitable responses by ranking candidate responses. But there is an obvious drawback for these methods that the responses are selected from a fixed response set and it is not possible to produce new responses for special inputs. SMT based methods treat response generation as a SMT problem on post-response parallel data. These methods are purely data-driven and can generate new responses. 

More recently, neural network based methods are being applied in this field \cite{serban2015building,yao2015attention,
	li-EtAl:2016:EMNLP20162}. In particular, \textsc{Seq2Seq} model and reinforcement learning are used to improve the quality of generated responses \cite{li-EtAl:2016:EMNLP20162}. Adversarial learning are also applied in this field in recent years \cite{li-EtAl:2017:EMNLP20175}. 
	\cite{serban2017hierarchical} introduced stochastic latent variable into RNN model into the response generation problem.
	Neural network based methods are promising for dialogue generation. However, as mentioned in section 2, training a neural network model requires a large corpus. Sometimes it is hard to obtain a large corpus in a specific domain, which limits their performance.

Another kind of related work is computational humor. Humor recognition or computation in natural language is still a challenging task. Although understanding universal humor characteristics is almost impossible, there are many attempts to capture latent structure behind humor. Taylor \shortcite{taylor2009computational} used ontological semantics to detect humor. Yang \shortcite{yang2015humor} identified several semantic structures behind humor and employed a computational approach to recognizing humor. Other studies also investigate humor with spoken or multimodal signals \cite{purandare2006humor}. But none of these works provide a systematical explanation of humor, not to mention recognizing humor in Chinese crosstalks. 

Moreover, there are several studies attempting to generate puns and jokes. For example, The JAPE system was developed to automatically generate punning riddles \cite{binsted1994implemented,binsted1997computational}, and it relies on a template-based NLG system, combining fixed text with slots. Following the seminal work of Binsted and Ritchie, the HAHAcronym system was developed to produce humorous acronyms \cite{stock2005act} and the subsequent system of \cite{binsted2003pun} focuses on the generation of referential jokes. More recently, an interesting unsupervised alternative to this earlier work was offered \cite{petrovic2013unsupervised}, and it does not require labeled examples or hard-coded rules. It starts from a template involving three slots and then finds funny triples. However, the task of entertaining dialogue generation has not been investigated. 



%


\section{Conclusions and Future Work}

In this paper, we investigate the possibility of automatic generation of entertaining dialogues in Chinese crosstalks. We proposed a humor-enhanced translation model to generate the replying utterance of the supporting role, given the utterance of the leading comedian in Chinese crosstalks. Evaluation results on a real Chinese crosstalk dataset verify the efficacy of our proposed model, especially the usefulness of the humor model.  

In future work, we will try to enlarge the dataset by exploiting dialogue data in other similar domains, aiming at further improving the performance. We will also investigate generating the utterance of the leading role in the crosstalks, given the context utterances in several previous turns of dialogues. 

\bibliography{aaai18}

\begin{thebibliography}{}

\bibitem[\protect\citeauthoryear{Banchs and Li}{2012}]{banchs2012iris}
Banchs, R.~E., and Li, H.
\newblock 2012.
\newblock Iris: a chat-oriented dialogue system based on the vector space
  model.
\newblock In {\em ACL},  37--42.
\newblock ACL.

\bibitem[\protect\citeauthoryear{Binsted and
  Ritchie}{1994}]{binsted1994implemented}
Binsted, K., and Ritchie, G.
\newblock 1994.
\newblock An implemented model of punning riddles.
\newblock Technical report, University of Edinburgh, Department of Artificial
  Intelligence.

\bibitem[\protect\citeauthoryear{Binsted and
  Ritchie}{1997}]{binsted1997computational}
Binsted, K., and Ritchie, G.
\newblock 1997.
\newblock Computational rules for generating punning riddles.
\newblock {\em HUMOR-International Journal of Humor Research} 10(1):25--76.

\bibitem[\protect\citeauthoryear{Binsted, Bergen, and
  McKay}{2003}]{binsted2003pun}
Binsted, K.; Bergen, B.; and McKay, J.
\newblock 2003.
\newblock Pun and non-pun humour in second-language learning.
\newblock In {\em Workshop Proceedings, CHI}.

\bibitem[\protect\citeauthoryear{Huang and
  Hsieh}{2010}]{huang2010infrastructure}
Huang, C.-R., and Hsieh, S.-K.
\newblock 2010.
\newblock Infrastructure for cross-lingual knowledge representation-towards
  multilingualism in linguistic studies.
\newblock {\em Taiwan NSC-granteLDBd Research Project (NSC
  96-2411-H-003-061-MY3)}.

\bibitem[\protect\citeauthoryear{Ji, Lu, and Li}{2014}]{ji2014information}
Ji, Z.; Lu, Z.; and Li, H.
\newblock 2014.
\newblock An information retrieval approach to short text conversation.
\newblock {\em arXiv preprint arXiv:1408.6988}.

\bibitem[\protect\citeauthoryear{Koehn \bgroup et al\mbox.\egroup
  }{2007}]{koehn2007moses}
Koehn, P.; Hoang, H.; Birch, A.; Callison-Burch, C.; Federico, M.; Bertoldi,
  N.; Cowan, B.; Shen, W.; Moran, C.; Zens, R.; et~al.
\newblock 2007.
\newblock Moses: Open source toolkit for statistical machine translation.
\newblock In {\em ACL},  177--180.
\newblock ACL.

\bibitem[\protect\citeauthoryear{Koehn, Och, and
  Marcu}{2003}]{koehn2003statistical}
Koehn, P.; Och, F.~J.; and Marcu, D.
\newblock 2003.
\newblock Statistical phrase-based translation.
\newblock In {\em NAACL HLT},  48--54.
\newblock ACL.

\bibitem[\protect\citeauthoryear{Lefcourt}{2001}]{lefcourt2001humor}
Lefcourt, H.~M.
\newblock 2001.
\newblock {\em Humor: The psychology of living buoyantly}.
\newblock Springer Science \& Business Media.

\bibitem[\protect\citeauthoryear{Levin, Pieraccini, and
  Eckert}{2000}]{levin2000stochastic}
Levin, E.; Pieraccini, R.; and Eckert, W.
\newblock 2000.
\newblock A stochastic model of human-machine interaction for learning dialog
  strategies.
\newblock {\em IEEE Transactions on speech and audio processing} 8(1):11--23.

\bibitem[\protect\citeauthoryear{Li \bgroup et al\mbox.\egroup
  }{2016}]{li-EtAl:2016:EMNLP20162}
Li, J.; Monroe, W.; Ritter, A.; Jurafsky, D.; Galley, M.; and Gao, J.
\newblock 2016.
\newblock Deep reinforcement learning for dialogue generation.
\newblock In {\em EMNLP}.

\bibitem[\protect\citeauthoryear{Li \bgroup et al\mbox.\egroup
  }{2017}]{li-EtAl:2017:EMNLP20175}
Li, J.; Monroe, W.; Shi, T.; Jean, S.; Ritter, A.; and Jurafsky, D.
\newblock 2017.
\newblock Adversarial learning for neural dialogue generation.
\newblock In {\em Proceedings of the 2017 Conference on Empirical Methods in
  Natural Language Processing},  2147--2159.
\newblock Copenhagen, Denmark: Association for Computational Linguistics.

\bibitem[\protect\citeauthoryear{Liaw and Wiener}{2002}]{Liaw2002}
Liaw, A., and Wiener, M.
\newblock 2002.
\newblock Classification and regression by randomforest.
\newblock {\em R News} 2(3):18--22.

\bibitem[\protect\citeauthoryear{Link}{1979}]{link1979genie}
Link, E.~P.
\newblock 1979.
\newblock {\em The genie and the lamp: Revolutionary Xiangsheng}.
\newblock publisher not identified.

\bibitem[\protect\citeauthoryear{Mackerras}{2013}]{mackerras2013performing}
Mackerras, C.
\newblock 2013.
\newblock {\em The performing arts in contemporary China}, volume~18.
\newblock Routledge.

\bibitem[\protect\citeauthoryear{Mihalcea and
  Strapparava}{2005}]{mihalcea2005making}
Mihalcea, R., and Strapparava, C.
\newblock 2005.
\newblock Making computers laugh: Investigations in automatic humor
  recognition.
\newblock In {\em HLT/EMNLP},  531--538.
\newblock ACL.

\bibitem[\protect\citeauthoryear{Moser}{1990}]{moser1990reflexivity}
Moser, D.
\newblock 1990.
\newblock Reflexivity in the humor of xiangsheng.
\newblock {\em CHINOPERL} 15(1):45--68.

\bibitem[\protect\citeauthoryear{Och}{2003}]{Och:2003:MER:1075096.1075117}
Och, F.~J.
\newblock 2003.
\newblock Minimum error rate training in statistical machine translation.
\newblock In {\em ACL}, ACL '03,  160--167.
\newblock ACL.

\bibitem[\protect\citeauthoryear{Oh and
  Rudnicky}{2000}]{Oh:2000:SLG:1117562.1117568}
Oh, A.~H., and Rudnicky, A.~I.
\newblock 2000.
\newblock Stochastic language generation for spoken dialogue systems.
\newblock In {\em ANLP/NAACL-ConvSyst},  27--32.
\newblock ACL.

\bibitem[\protect\citeauthoryear{Papineni \bgroup et al\mbox.\egroup
  }{2002}]{papineni2002bleu}
Papineni, K.; Roukos, S.; Ward, T.; and Zhu, W.-J.
\newblock 2002.
\newblock Bleu: a method for automatic evaluation of machine translation.
\newblock In {\em ACL},  311--318.
\newblock ACL.

\bibitem[\protect\citeauthoryear{Paulos}{2008}]{paulos2008mathematics}
Paulos, J.~A.
\newblock 2008.
\newblock {\em Mathematics and humor: A study of the logic of humor}.
\newblock University of Chicago Press.

\bibitem[\protect\citeauthoryear{Petrovic and
  Matthews}{2013}]{petrovic2013unsupervised}
Petrovic, S., and Matthews, D.
\newblock 2013.
\newblock Unsupervised joke generation from big data.
\newblock In {\em ACL (2)},  228--232.
\newblock Citeseer.

\bibitem[\protect\citeauthoryear{Pieraccini \bgroup et al\mbox.\egroup
  }{2009}]{pieraccini2009we}
Pieraccini, R.; Suendermann, D.; Dayanidhi, K.; and Liscombe, J.
\newblock 2009.
\newblock Are we there yet? research in commercial spoken dialog systems.
\newblock In {\em TSD},  3--13.
\newblock Springer.

\bibitem[\protect\citeauthoryear{Purandare and
  Litman}{2006}]{purandare2006humor}
Purandare, A., and Litman, D.
\newblock 2006.
\newblock Humor: Prosody analysis and automatic recognition for f*r*i*e*n*d*s.
\newblock In {\em EMNLP},  208--215.
\newblock ACL.

\bibitem[\protect\citeauthoryear{Ritter, Cherry, and
  Dolan}{2011}]{ritter2011data}
Ritter, A.; Cherry, C.; and Dolan, W.~B.
\newblock 2011.
\newblock Data-driven response generation in social media.
\newblock In {\em EMNLP},  583--593.
\newblock ACL.

\bibitem[\protect\citeauthoryear{Serban \bgroup et al\mbox.\egroup
  }{2015}]{serban2015building}
Serban, I.~V.; Sordoni, A.; Bengio, Y.; Courville, A.; and Pineau, J.
\newblock 2015.
\newblock Building end-to-end dialogue systems using generative hierarchical
  neural network models.
\newblock {\em arXiv preprint arXiv:1507.04808}.

\bibitem[\protect\citeauthoryear{Serban \bgroup et al\mbox.\egroup
  }{2017}]{serban2017hierarchical}
Serban, I.~V.; Sordoni, A.; Lowe, R.; Charlin, L.; Pineau, J.; Courville,
  A.~C.; and Bengio, Y.
\newblock 2017.
\newblock A hierarchical latent variable encoder-decoder model for generating
  dialogues.
\newblock In {\em AAAI},  3295--3301.

\bibitem[\protect\citeauthoryear{Sordoni \bgroup et al\mbox.\egroup
  }{2015}]{sordoni-EtAl:2015:NAACL-HLT}
Sordoni, A.; Galley, M.; Auli, M.; Brockett, C.; Ji, Y.; Mitchell, M.; Nie,
  J.-Y.; Gao, J.; and Dolan, B.
\newblock 2015.
\newblock A neural network approach to context-sensitive generation of
  conversational responses.
\newblock In {\em NAACL HLT}.

\bibitem[\protect\citeauthoryear{Stock and Strapparava}{2005}]{stock2005act}
Stock, O., and Strapparava, C.
\newblock 2005.
\newblock The act of creating humorous acronyms.
\newblock {\em Applied Artificial Intelligence} 19(2):137--151.

\bibitem[\protect\citeauthoryear{Taylor}{2009}]{taylor2009computational}
Taylor, J.~M.
\newblock 2009.
\newblock Computational detection of humor: A dream or a nightmare? the
  ontological semantics approach.
\newblock In {\em WI-IAT},  429--432.
\newblock IEEE Computer Society.

\bibitem[\protect\citeauthoryear{Terence}{2013}]{terence2013china}
Terence, H.
\newblock 2013.
\newblock China's comedy showdown.
\newblock {\em The World of Chinese} 3(2):48--51.

\bibitem[\protect\citeauthoryear{Xu \bgroup et al\mbox.\egroup }{2008}]{Xu2008}
Xu, H.; Lin, H.; Pan, Y.; Hui, R.; and Chen, j.
\newblock 2008.
\newblock Constructing the affective lexicon ontology.
\newblock {\em Journal of the China Society for Scientific and Technical
  Information} 27(2):180--185.

\bibitem[\protect\citeauthoryear{Yang \bgroup et al\mbox.\egroup
  }{2015}]{yang2015humor}
Yang, D.; Lavie, A.; Dyer, C.; and Hovy, E.~H.
\newblock 2015.
\newblock Humor recognition and humor anchor extraction.
\newblock In {\em EMNLP},  2367--2376.
\newblock ACL.

\bibitem[\protect\citeauthoryear{Yao, Zweig, and Peng}{2015}]{yao2015attention}
Yao, K.; Zweig, G.; and Peng, B.
\newblock 2015.
\newblock Attention with intention for a neural network conversation model.
\newblock {\em arXiv preprint arXiv:1510.08565}.

\bibitem[\protect\citeauthoryear{Zaidan}{2009}]{zaidan2009z}
Zaidan, O.
\newblock 2009.
\newblock Z-mert: A fully configurable open source tool for minimum error rate
  training of machine translation systems.
\newblock {\em The Prague Bulletin of Mathematical Linguistics} 91:79--88.

\bibitem[\protect\citeauthoryear{Zhang and Liu}{2014}]{zhang2014recognizing}
Zhang, R., and Liu, N.
\newblock 2014.
\newblock Recognizing humor on twitter.
\newblock In {\em CIKM},  889--898.
\newblock ACM.

\end{thebibliography}
\bibliographystyle{aaai}
\end{CJK*}
\end{document}